\documentclass[sn-mathphys-num]{sn-jnl}

\usepackage{amsthm}%
\usepackage{algorithm}%
\usepackage{algorithmicx}%
\usepackage{algpseudocode}%
\usepackage[title]{appendix}%
\usepackage{amsmath,amssymb,amsfonts}%
\usepackage{bm}%
\usepackage{booktabs}%
\usepackage{color}%
\usepackage{caption}%
\usepackage{colortbl}%
\usepackage{graphicx}%
\usepackage{listings}%
\usepackage{multirow}%
\usepackage{mathrsfs}%
\usepackage{manyfoot}%
\usepackage[numbers,sort&compress]{natbib}%
\usepackage{soul}%
\usepackage{subcaption}%
\usepackage{threeparttable}%
\usepackage{textcomp}%
\usepackage{xcolor}%

\newcommand{\tablestyle}[2]{\setlength{\tabcolsep}{#1}\renewcommand{\arraystretch}{#2}\centering\footnotesize}

\definecolor{tablered}{RGB}{205,51,51}
\definecolor{tablegreen}{HTML}{39b54a}
\definecolor{tableblue}{HTML}{4682B4}

\sethlcolor{yellow}
\soulregister{\cite}7
\soulregister{\ref}7
\soulregister{\textbf}7
\soulregister{\textit}7
\soulregister{\emph}7

\theoremstyle{thmstyleone}%
\theoremstyle{thmstyletwo}%

\theoremstyle{thmstylethree}%

\begin{document}

\title[Article Title]{Catheter Detection and Segmentation in X-ray Images via Multi-task Learning}


\author*[1]{\fnm{Lin} \sur{Xi}}\email{l.xi@uea.ac.uk}

\author*[1,2]{\fnm{Yingliang} \sur{Ma}}\email{yingliang.ma@uea.ac.uk}

\author[1]{\fnm{Ethan} \sur{Koland}}\email{e.koland@uea.ac.uk}

\author[2]{\fnm{Sandra} \sur{Howell}}\email{sandra.howell@kcl.ac.uk}

\author[2]{\fnm{Aldo} \sur{Rinaldi}}\email{aldo.rinaldi@kcl.ac.uk}

\author[2]{\fnm{Kawal S.} \sur{Rhode}}\email{kawal.rhode@kcl.ac.uk}

\affil*[1]{\orgdiv{School of Computing Sciences}, \orgname{University of East Anglia}, \orgaddress{\city{Norwich}, \postcode{NR4 7TJ}, \country{United Kingdom}}}

\affil[2]{\orgdiv{School of Biomedical Engineering and Imaging Sciences}, \orgname{King's College London}, \orgaddress{\city{London}, \postcode{SE1 7EH}, \country{United Kingdom}}}

\abstract{
	\textbf{Purpose} Automated detection and segmentation of surgical devices, such as catheters or wires, in X-ray fluoroscopic images have the potential to enhance image guidance in minimally invasive heart surgeries.
	
	\textbf{Methods} In this paper, we present a convolutional neural network model that integrates a resnet architecture with multiple prediction heads to achieve real-time, accurate localization of electrodes on catheters and catheter segmentation in an end-to-end deep learning framework. We also propose a multi-task learning strategy in which our model is trained to perform both accurate electrode detection and catheter segmentation simultaneously. A key challenge with this approach is achieving optimal performance for both tasks. To address this, we introduce a novel multi-level dynamic resource prioritization method. This method dynamically adjusts sample and task weights during training to effectively prioritize more challenging tasks, where task difficulty is inversely proportional to performance and evolves throughout the training process.
	
	\textbf{Results} Experiments on both public and private datasets have demonstrated that the accuracy of our method surpasses the existing state-of-the-art methods in both single segmentation task and in the detection and segmentation multi-task.
	
	\textbf{Conclusion} Our approach achieves a good trade-off between accuracy and efficiency, making it well-suited for real-time surgical guidance applications.
}


\maketitle

\section{Introduction}\label{sec:intro}

Minimally invasive heart surgeries are routinely carried out to treat heart diseases such as atrial fibrillation (AF), heart failure, congenital heart diseases, and more. The surgery is usually guided using X-ray fluoroscopy. Catheters and pacing leads in the form of wires are used as surgical devices to carry out the treatment and they are highly visible in X-ray images. However, heart chambers and blood vessels are hardly visible under X-ray. To enhance the procedure guidance, 3D models of heart chambers and blood vessels extracted from pre-procedural CT or MR scans can be overlaid on the top of X-ray images to add anatomical information \cite{knecht2008computed}. To improve the accuracy of the enhanced guidance, motion compensation is added to the 3D models to allow them moving together with the patient's cardiac and respiratory motions by tracking a stationary catheter or wire in X-ray images \cite{Ma2012TBME}. Furthermore, detecting catheters in X-ray images enables the automatic registration between 3D heart models and 2D X-ray images \cite{Truong2017STACOM}. Finally, knowing the locations of catheters and wires may allow procedures with complete or shared autonomy with robots in the near future. For catheter detection, early work is focused on active contours and shape models \cite{Schenderlein2010MI, BROST2010MIA}. Recently, vessel enhancement filters were used to extract the body of catheters and identify the type of catheter \cite{Ma2018MP, Ma2022TBME}. However, these methods are prone to errors due to image artifacts and the presence of other wire-like objects. Learning-based approaches have been developed to build a shape template to continuously track guidewires \cite{Ambrosini2017MICCAI, Wu2011CVPR}. They use manual feature extraction, and the methods are less robust when the image contains numerous wire-like structures and they only track one particular object.

Implementing a real-time unified model for both detection and segmentation tasks is essential for achieving a more accurate and streamlined integration approach in robotic systems. The most accurate two-stage object detection methods \cite{FastRCNN, FasterRCNN} are not suitable for developing a unified model, and incorporating a one-stage object detection \cite{YOLO, DETR} with a segmentation model \cite{FCN, DeepLabV3} is challenging to achieve real-time speed. Our main goals are to determine the catheter location and region through more accurate and real-time catheter detection and segmentation algorithms, and then use post-processing algorithms to localize the electrode positions and wire centerline of the catheter.

The rise of deep learning (DL) for surgery instruments (i.e., catheter \cite{Ranne2024ICRA}, ultrasound probes \cite{Ma2024CIP, Ma2021PMB}) detection and segmentation in X-ray images offers an opportunity for the development of robust catheter detection methods. DL-based methods, similar to humans, can perform multi-task learning by training on multiple tasks simultaneously, such as classification, detection, and segmentation \cite{Chen2023TCBB, Nguyen2020ICRA}. Inspired by human learning processes \cite{Bellotti2004}, DL models allocate resources based on the complexity and difficulty of each sample and task, thereby improving the effectiveness and efficiency of the learning process \cite{ELMAN1993C, yang2016deep, jou2016deep}. Curriculum learning (CL) focuses on learning easy tasks first and harder ones later, distinguishing between basic and advanced tasks \cite{ELMAN1993C}. In the CL approach \cite{Bengio2009ICML}, tasks are broken down into subtasks that follow an easy-to-hard training strategy. However, CL can struggle in multi-task problems, as it assumes a consistent underlying distribution across tasks, which may not hold when dealing with different types, like segmentation and classification in medical imaging. Dynamic task prioritization (DTP) addresses this by encouraging the model to focus on difficult tasks \cite{Guo2018ECCV} and embedding explicit task priorities into the neural network.

In this paper, we propose a multi-task framework with multi-level dynamic resource prioritization for medical image analysis. Inspired by CL \cite{ELMAN1993C, Bengio2009ICML}, our model is designed to prioritize difficult samples and tasks simultaneously, similar to the optimization of hard negative examples \cite{Xuan2020ECCV, Robinson2021ICLR} across different samples and tasks. This approach results in more generalizable features and improves object detection and segmentation, achieving performance that surpasses state-of-the-art (SoTA) methods on both public and private datasets. Our contributions are: 

\begin{itemize}
	\item A novel convolutional neural network architecture capable of object detection using a center heatmap and object segmentation.
	\item A novel multi-level dynamic resource prioritization training strategy for multi-task learning. It effectively allocates learning resources to difficult samples/tasks rather than easier samples/tasks at both a sample-level and task-level.
\end{itemize}

\section{Methods}\label{sec:met}

\subsection{Network Formulation}\label{subsec:met:network}

\begin{figure*}[!htbp]
	\begin{center}
		\includegraphics[width=0.8\textwidth]{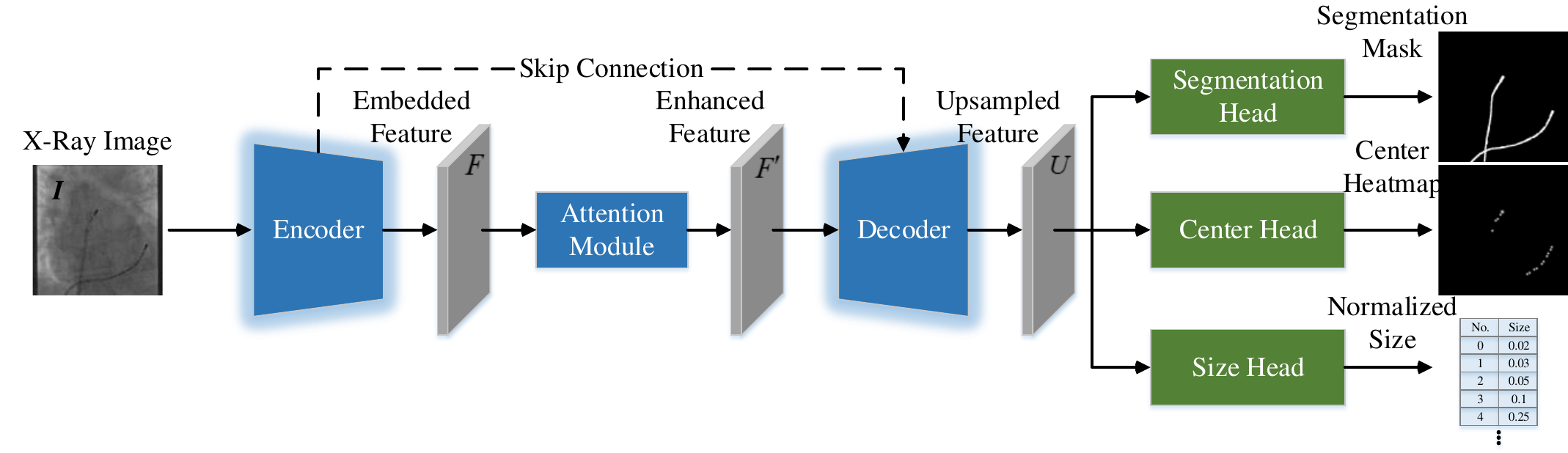}
	\end{center}
	\caption{The overview of our model. Given an X-ray image $\bm{I}$, we utilize an encoder to embed it into a 512-dimensional embedding feature $\bm{F}$ and then fed into an attention module to produce enhanced feature $\bm{F}'$. The enhanced features $\bm{F}'$ further fed into a decoder to recover resolution via a top-down manner with a skip connection. Finally, the output of the top-down decoder is passed to segmentation, center, and size prediction heads to obtain final results.}
	\label{fig:framework}
\end{figure*}

The model is designed in an encoder-decoder fashion since this kind of architecture is able to preserve low-level details to refine high-level global contexts. More specifically, our model comprises three key modules: encoder, attention module, and decoder. The overall framework of the model is shown in Fig. \ref{fig:framework}.

The encoder takes the X-Ray image $\bm{I}$ as the input. The encoder outputs an embedded feature map $\bm{F}$ attached to the backbone network. We take \texttt{res4} features with stride 16 from the base ResNets as our backbone features and discard \texttt{res5}. A 3$\times$3 convolutional layer without non-linearity is used as a projection head from the backbone feature to the latent feature space. We set latent feature dimensions to 512. The output of the embedding is a 2D map ($\bm{F}\in\mathbb{R}^{H\times W\times C}$), where H is the height, W is the width, and C is the feature dimension of the backbone network output feature map.

Similar to attention map inference along two separate dimensions of CBAM \cite{CBAM}, namely channel and spatial, we add an attention module $A(\cdot)$ to apply temporal and spatial fusion for the embedding feature map $\bm{F}$. The enhanced features $\bm{F}'$ are fed into a decoder. Features are processed and upsampled at a scale of two gradually with higher-resolution features from the attention module incorporated using skip-connections with an encoder. The final layer of the decoder produces a stride 4 feature $\bm{U}$ which is fed into the center, size, and segmentation heads and bilinearly upsampled to the original resolution.

\subsection{Multi-Level Dynamic Resource Prioritization}

We introduce a multi-level dynamic resource prioritization strategy for biomedical image analysis. Unlike the DTP in \cite{Guo2018ECCV}, which assigns different weights by the focal loss-like weighting strategy at the task level, our method directly adjusts the weights based on the performance metric across both samples and tasks. Additionally, unlike \cite{Li2017AAAI}, our approach does not rely on task losses to determine the relative difficulty of tasks. Instead, we use more intuitive and realistic metrics for dynamically prioritizing tasks: task performance --- also known as key performance metrics (KPIs) in \cite{Guo2018ECCV}. To reasonably arrange resources, we define the notion of priority and discuss how we dynamically adjust it, based on training difficulty. There are two use cases: (i) sample-level priority and (ii) task-level priority.

\noindent \textbf{Key Performance Indicators.} For each task $T_{t}\ (t\in\{\texttt{d},\texttt{s}\})$, we select a key performance indicator (KPI) denoted by $\kappa_{t}\in[0, 1]$ and should be a meaningful metric such as accuracy or average precision, where $\texttt{d}$ and $\texttt{s}$ represent the detection and segmentation task, respectively. For each task, we use mean absolute error (MAE) or root mean square error (RMSE) for the detection task and dice similarity coefficient (DSC) or intersection-over-union (IoU) for the segmentation task in the catheter dataset. We also define difficulty $\mathcal{D}\propto\kappa^{-1}$ to sort samples and tasks ordered by the difficulty $\mathcal{D}$.

In training process, we update $\kappa$ to be an exponential moving average $\bar{\kappa}^{(\tau)}_{t}=\alpha\kappa^{(\tau)}_{t}+(1-\alpha)\bar{\kappa}^{(\tau-1)}_{t}$ where $\tau$ is the training iteration number and $\alpha\in[0, 1]$ is the discount factor. Larger values of $\alpha$ prioritize more recent examples.

\noindent \textbf{Sample-Level Prioritization.} For task $T_{t}$, we first define the task-specific loss (\emph{e.g.}, cross-entropy) denoted by $L_{t}(\cdot)$. Since some samples may not be available for specific tasks $T_{t}$ at a particular training time, we use $\delta_{t, i}\in{0, 1}$ to denote the availability of ground-truth data for sample $i$, task $T_{t}$. Then we use $\delta_{t, i}$ to mask task loss $\mathcal{L}_{t}(\cdot)$ is defined in
\begin{equation}\label{eq:1}
	\mathcal{L}_{t}(\cdot)=\frac{1}{N}\sum_{i=1}^{N}\delta_{t, i}L_{t}(p_{t}^{i},y_{t}^{i}),
\end{equation}
where $i$ is the index of the training sample, $p_{t}^{i}$ is the model's output for sample $i$ for task $T_{t}$, and $y_{t}^{i}$ is the ground-truth for sample $i$ for task $T_{t}$.

We now describe how difficult samples are identified. Consider pixel-wise binary classification with cross-entropy (CE):
\begin{equation}\label{eq:2}
	\mathrm{CE}(j)=-\log(\mathtt{softmax}(p^{(j)})),
\end{equation}
where $p^{(j)}$ is the model segmentation logit result for the $j$-th pixel.

We down-weight easier samples and focus on harder samples during training based on sample performance metrics. Let $i$ denote the current sample index being considered from the dataset. Samples are ordered according to their difficulty $\mathcal{D}_{S}(i)$. We assign different priorities to samples by using sample-level weights. After that, we solved masked weight $\delta_{t, i}$ by performing a hard assignment of the sample-level weights which is determined by the sample difficulty $\mathcal{D}_{S}(i)$ with a threshold of $\eta$. Unlike \cite{Guo2018ECCV}, our weight parameters are discrete $\delta_{t, i}$ and defined by the threshold after top-k filtering \cite{Cheng2021CVPR}. In summary, the $\delta_{t, i}$ for sample $i$, task $t$ can be computed by:
\begin{equation}\label{eq:3}
	\delta_{t, i}=\left\{\begin{matrix}
		0,&\ \textrm{if}\ i\notin \mathtt{Top}^{K}(\mathcal{D}_{S}(i))\\1,&\ \textrm{otherwise}
	\end{matrix}\right.,
\end{equation}
where $\mathtt{Top}^{K}(\mathcal{D}_{S}(i))$ denotes the set of indices that are top-k filtering in the dataset.

\noindent \textbf{Task-Level Prioritization.} Similar to sample-level prioritization, we use task-specific KPI to assign weight to tasks. If the KPI$\ll$0.5, we can assume that task $T_{t}$ is difficult for the model and this should be assigned more resources for training. To balance easy and difficult tasks, we proposed to scale each task-specific loss $\mathcal{L}_{t}$ by computing the task difficulty $\mathcal{D}_{T}(t)=1/\bar{\kappa}_{t}$. Our dynamic resource prioritization loss $\mathcal{L}_{total}$ is:
\begin{equation}\label{eq:4}
	\mathcal{L}_{total}=\frac{1}{\bar{\kappa}_{t}}\mathcal{L}_{t}(\cdot),
\end{equation}

To summarize, our total loss $\mathcal{L}_{total}$ uses learning progress signals (\emph{i.e.}, $\bar{\kappa}_{t}$) to automatically compute a priority level at both a task-level and sample-level. These priority levels vary throughout the training procedure. On the other hand, the DTP \cite{Guo2018ECCV} is a focal loss-based approach with focusing parameters to sort the tasks and is easily influenced by hyperparameters.

For catheter detection and segmentation task, we define $\kappa_{t}$ as follow:
\begin{equation}\label{eq:5}
	\kappa_{\texttt{d}}=\textrm{MAE}(\mathcal{S}_\texttt{d})^{-1}\ \ \kappa_{s}=\textrm{IoU}(\mathcal{S}_\texttt{s}),
\end{equation}
where $\mathcal{S}_\texttt{d}$ and $\mathcal{S}_\texttt{s}$ are training samples for detection and segmentation tasks.

\noindent \textbf{Loss Functions} There were two loss functions for multi-task learning for the detection and segmentation of catheters in X-ray images. That is, the final loss is
\begin{equation}\label{eq:6}
	\mathcal{L}_{\textrm{total}}=\frac{1}{\bar{\kappa}_{\texttt{d}}}\mathcal{L}_{\texttt{d}}+\frac{1}{\bar{\kappa}_{\texttt{s}}}\mathcal{L}_{\texttt{s}}
\end{equation}

For the catheter detection task, following \cite{Zhou2019ArXiv}, the loss function for catheter heatmap regression training is focal loss, and we use an L1 loss at the box size,
\begin{equation}\label{eq:7}
	\mathcal{L}_{\texttt{d}}=\mathcal{L}_{focal}+\lambda\mathcal{L}_{size}
\end{equation}
where $\lambda$ is set to 1.

For the segmentation task, we adopt the same loss function with \cite{BASNet} to jointly measure the prediction at the pixel level by binary cross entropy (BCE) loss as well as in the region level by IoU loss:
\begin{equation}\label{eq:lf:1}
	\mathcal{L}_{\texttt{s}}(\hat{M}, M)=\mathcal{L}_{bce}(\hat{M}, M)+\mathcal{L}_{iou}(\hat{M}, M),
\end{equation}
where $\hat{M}$ denotes the segmentation prediction and $M$ refers to the binary ground-truth.

\section{Experiments}\label{sec:exp}

\subsection{Experimental Setup}

\noindent\textbf{Datasets and evaluation metrics}. To train and test our model, we use a public dataset (UCL catheter segmentation dataset \cite{Gherardini2020CMPB}, 3000 images) and a private dataset. The private dataset contains 2450 X-ray images, which were acquired in 62 different clinical cases using two mono-plane X-ray systems (both are Philips X-ray systems) at St. Thomas' Hospital London and University Hospitals Coventry \& Warwickshire. All clinical cases are standard atrial fibrillation ablation procedures. The manual labeling of catheters and wires in X-ray images is very time-consuming and tedious. To speed the process up, vessel enhancement filters \cite{Frangi1998MICCAI} were used to extract catheters and wires. The resulting image was automatically binarized by an adaptive binarization method \cite{Ma2018MP}. Not all wires were labeled in our training data. As we are only interested in surgical devices inside the heart, stationary wires such as ECG leads and sternal wires from open-heart surgeries are not labeled. Therefore, an experienced clinician manually removed the non-target objects.

To test the performance of our multi-task learning strategy, we carry out comprehensive experiments and evaluate the model performance by object detection precision and region similarity. For the segmentation evaluation, we adopt a standard metric suggested by \cite{DAVIS2016}, namely region similarity $\mathcal{J}$, which is the IoU of the prediction and the ground-truth. In the detection task, we report the average precision over all IoU thresholds (AP).

\noindent\textbf{Implementation details.} The backbone of the model is ResNet34 \cite{ResNet}, for each input image of size $256\times 256 \times 1$, the image is down-sampled to the size of $\left\{128, 64, 32, 16\right\}$ in the first four layers of ResNet34. The channel dimension of the attention module is set to 512. For the prediction, we add a separate head for each prediction including segmentation, center, and size prediction heads, respectively. We implement each prediction head by using two 3$\times$3 convolutional layers with 64 channels and a final 1$\times$1 convolution then produce the desired output.

The whole network is trained using the AdamW optimizer ($\beta_{1}$=0.9 and $\beta_{2}$=0.999) with a learning rate of $10^{-4}$. We first initialize our model by pre-training for 100 epochs each on the Duke OCT \cite{Srinivasan2014BOE} and UCL catheter segmentation \cite{Gherardini2020CMPB} datasets. Then, we train our network for 100 epochs on our private dataset in the multi-task object detection and segmentation training stage using our proposed multi-level dynamic resource prioritization strategy. Data augmentation (\emph{e.g.}, scaling, flipping, and rotation) is also adopted for both images and video data. Our model is implemented in PyTorch. All experiments and analyses are conducted on an NVIDIA RTX 6000 GPU, and the overall training time is about 5 hours.

\subsection{Comparisons with SoTA}

We compare our proposed multi-task learning model with the SoTA methods. We show the detailed results in Table \ref{tab:SoTA}, with seven biomedical image analysis methods, \emph{e.g.}, U-Net \cite{UNET}, Attention U-Net \cite{AttnUNET}, U-Net++ \cite{UNETPLAS}, U-Net 3+ \cite{UNET3PLUS}, TransU-Net \cite{TRANSUNET}, MedT \cite{MEDT}, UNeXt \cite{UNEXT}, Y-Net \cite{YNET}, SwinU-Net \cite{SWINUNET}, CMU-Net \cite{CMUNET}, CMU-NeXt \cite{CMUNEXT}, SANet \cite{SANET}, TransFuse \cite{TRANSFUSE}, FCN \cite{FCN}, and DeepLabV3 \cite{DeepLabV3}, taken from the biomedical image analysis benchmark. Note that all methods are trained with the same dataset and the same training strategy, \emph{i.e.}, the proposed multi-task learning method.

\begin{table*}[!htbp]
	\centering
	\captionsetup{font=small}
	\caption{Quantitative results on the \texttt{val} and \texttt{test} set of catheter detection and segmentation benchmarks, using the average precision AP and region similarity $\mathcal{J}$. The best performance scores are highlighted in \textbf{bold}.}
	\label{tab:SoTA}%
	\begin{threeparttable}
		\resizebox{1.0\linewidth}{!}{
			\tablestyle{12pt}{1.02}
			\begin{tabular}{@{\hskip 2pt}r|c|c|cccc|c@{\hskip 2pt}}\toprule
				\multirow{4}[0]{*}{Methdos} & \multirow{2}[0]{*}{\#Param.} & UCL  & \multicolumn{4}{c|}{Catheter Detection and Segmentation} & \multirow{2}[0]{*}{Runtime} \\
				&       & \texttt{val}   & \multicolumn{2}{c}{\texttt{val}} & \multicolumn{2}{c|}{\texttt{test}} &  \\
				& \multirow{2}[0]{*}{M} & Segmentation & Detection & Segmentation & Detection & Segmentation & \multirow{2}[0]{*}{FPS $\downarrow$} \\
				&       & Mean $\mathcal{J}\uparrow$  & AP $\uparrow$   & Mean $\mathcal{J}\uparrow$ & AP $\uparrow$   & Mean $\mathcal{J}\uparrow$ &  \\\midrule
				U-Net \cite{UNET} & 34.52 & 76.45 & 81.77 & 62.90  & 80.27 & 60.39 & 65 \\
				Attention U-Net \cite{AttnUNET} & 34.87 & 76.21 & 83.08 & 63.52 & 82.53 & 61.28 &  59 \\
				U-Net++ \cite{UNETPLAS}  & 26.90  & 76.48 & 81.58 & 62.97 & 80.06 & 60.24 &  57 \\
				U-Net 3+  \cite{UNET3PLUS} & 26.97 & 76.33 & 81.64 & 62.71 & 80.71 & 60.15 & 20 \\
				TransU-Net \cite{TRANSUNET} & 105.32 & 76.56 & 82.89 & 63.02 & 81.43 & 60.87 & 26 \\
				MedT \cite{MEDT}   & \textbf{1.37}  & 75.13 & 80.07 & 61.88 & 78.56 & 58.72 & 15 \\
				UNeXt \cite{UNEXT} & 1.47  & 75.19 & 80.38 & 62.36 & 79.99 & 58.99 & 143 \\
				Y-Net \cite{YNET} & 7.46  & 76.49 & 81.83 & 63.13 & 80.39 & 60.44 & 84 \\
				SwinU-Net \cite{SWINUNET} & 27.14 & 74.05 & 76.13 & 59.45 & 75.86 & 56.48 & \textbf{265} \\
				CMU-Net \cite{CMUNET} & 49.93 & 76.46 & 81.01 & 63.21 & 80.11 & 60.69 & 41 \\
				CMU-NeXt \cite{CMUNEXT} & 3.14  & 76.53 & 80.67 & 63.19 & 79.47 & 60.53 & 174 \\
				SANet \cite{SANET} & 23.90  & 77.18 & 83.35 & 64.78 & 82.87 & 63.01 & 76 \\
				TransFuse \cite{TRANSFUSE} & 26.33 & 76.54 & 82.96 & 63.88 & 82.63 & 61.84 & 101 \\
				FCN$^{*}$ \cite{FCN}   & 29.20  & 76.26 & 79.93 & 62.64 & 77.55 & 59.16 & 41 \\
				DeepLabV3$^{*}$ \cite{DeepLabV3}  & 35.88 & 77.07 & 82.47 & 64.55 & 80.97 & 62.38 & 31 \\\midrule
				\textbf{Ours}$^{*}$  & 32.39 & \textbf{77.21} & \textbf{84.15} & \textbf{65.37} & \textbf{83.13} & \textbf{63.97} &  37 \\\bottomrule
			\end{tabular}%
		}
		\begin{tablenotes}
			\footnotesize
			\item[]$*$ is use ResNet34 as the backbone. The abbreviation `M' in the `\#Param.' cell represents a million.
		\end{tablenotes}
	\end{threeparttable}
\end{table*}%

We observe that our proposed model delivers competitive performance on both the \texttt{val} and \texttt{test} datasets, compared to other existing methods. As shown in Table \ref{tab:SoTA}, our method achieves the results of 84.15\% and 65.37\% in terms of AP and mean $\mathcal{J}$ on the \texttt{val} set as well as 85.13\% and 63.97\% on the \texttt{test} set. Compared to Attention U-Net \cite{AttnUNET} and SANet \cite{SANET}, which are the best-performing methods in U-Net and ResNet-based architectures, our method based on a lighter weight architecture gives performance gains of \textbf{1.07}\%, \textbf{0.80}\% and \textbf{1.85}\%, \textbf{0.59}\% on AP and mean $\mathcal{J}$ on the \texttt{val} set, respectively. The same results can be observed on the \texttt{test} set. The results indicate that our proposed multi-task learning strategy can effectively improve the overall performance in object detection and segmentation. For the single segmentation task, our method achieves the best performance among all of the methods in terms of mean $\mathcal{J}$. Compared to the second-best method SANet \cite{SANET}, our model achieves a gain of \textbf{0.03}\% in mean $\mathcal{J}$.

The primary advantage of our architecture lies in its use of the first four layers of ResNet34, combined with an attention module implemented through the temporal and spatial fusion of the embedding feature map. This design enhances spatial stability, increases the attention receptive field, and helps recover important information lost in regions with attention values close to zero in the feature map. Additionally, the resolution of the X-ray images in both UCL dataset and catheter detection and segmentation dataset is 256$\times$256. However, when images are downsampled across five layers of the encoder, the activated feature area for a slender foreground object, such as a catheter, is significantly reduced, making it challenging to maintain clear and consistent boundaries.

\begin{figure*}[!htbp]
	\begin{center}
		\includegraphics[width=1.0\textwidth]{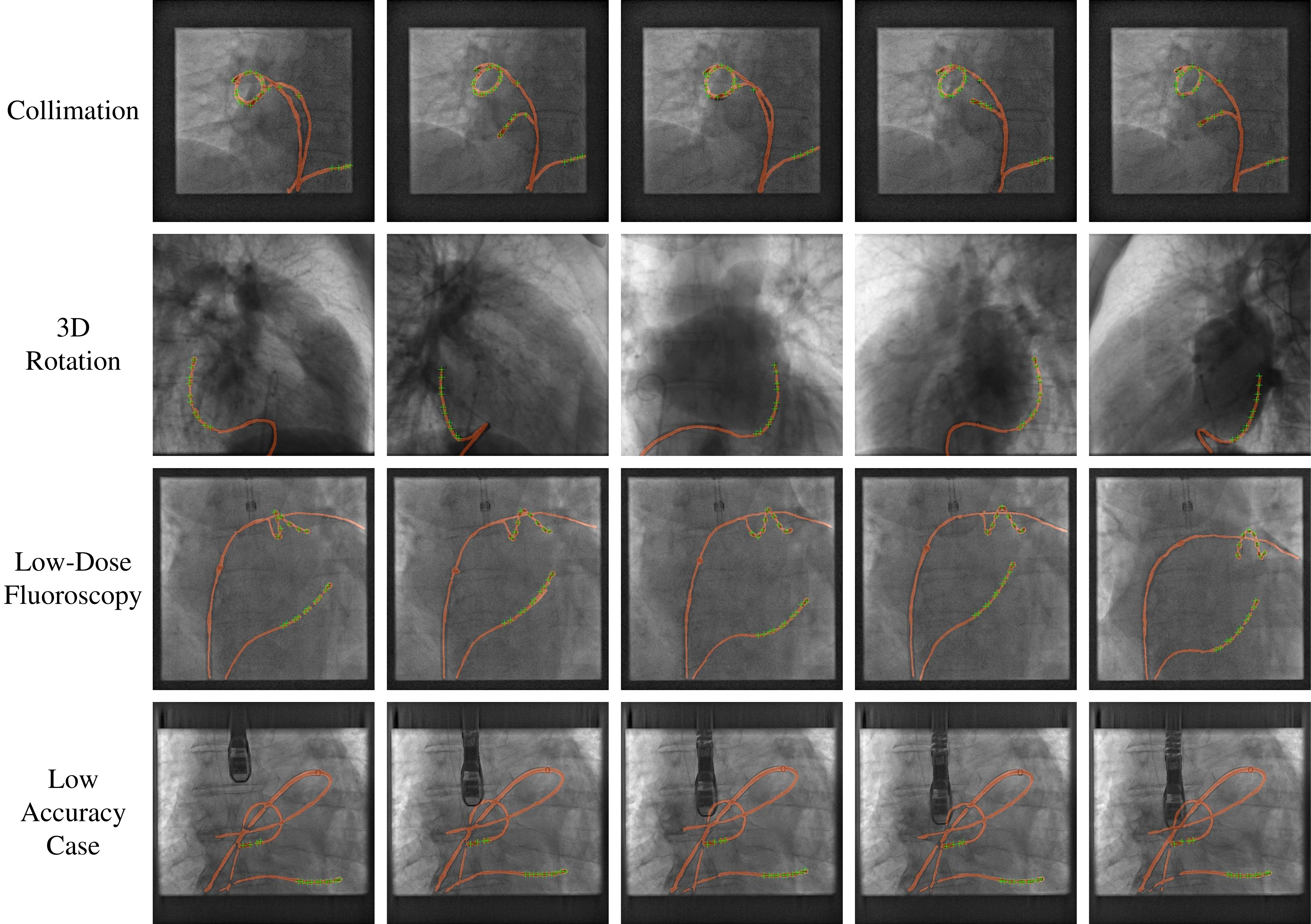}
	\end{center}
	\caption{Qualitative results of the proposed method on challenging scenarios from the catheter detection and segmentation dataset. The green crosses are the positions of electrodes. The orange mask indicates the segmentation results of the catheter.}
	\label{fig:ods_qual}
\end{figure*}

Fig. \ref{fig:ods_qual} visualizes our multi-task model catheter detection and segmentation results with the last row being a low accuracy case (Section \ref{sec:con}). We choose some X-ray sequences from the catheter detection and segmentation dataset with the cases of collimation, 3D rotation, and low-dose fluoroscopy. It can be seen that our model can effectively detect and locate the position of the catheter, and our method is able to discriminate the target catheter from complex background distractors.

\subsection{Ablation Study}

To demonstrate the influence of each component and hyper-parameters in our method, we perform an ablation study on the \texttt{val} set of our private catheter detection and segmentation dataset. The evaluation criterion is the object detection precision AP and the mean region similarity $\mathcal{J}$.

\noindent\textbf{Choice of Backbone.} For the backbone selection, we have compared the performance of the proposed method with different backbones, including VGG16 \cite{VGG}, ResNet18 \cite{ResNet}, ResNet34 \cite{ResNet}, HRNet-W18 \cite{HRNet}, and HRNet-W32 \cite{HRNet}, as shown in Table \ref{table:Abla:backbone}. To verify of the overall performance of the multi-task learning strategy, we also define a new evaluation criterion, the mean KPI which denotes the average value of AP and mean $\mathcal{J}$.

\begin{table*}[!htbp]
	\centering
	\captionsetup{font=small}
	\caption{Comparison of the several different backbones, measured by the AP, mean $\mathcal{J}$, Mean KPIs and FPS.}
	\label{table:Abla:backbone}
	\resizebox{0.8\linewidth}{!}{
		\tablestyle{12pt}{1.02}
		\begin{tabular}{@{\hskip 2pt}c|ccc|c@{\hskip 2pt}}\toprule
			Backbone  & AP $\uparrow$   & Mean $\mathcal{J}\uparrow$ & Mean KPI $\uparrow$ & FPS $\uparrow$ \\\midrule
			VGG16 \cite{VGG}     & 82.97 		   & 62.70  		& 72.84    		 & \textbf{57} \\
			ResNet18 \cite{ResNet} & 83.74 		   & 63.92  		& 73.83    		 & 45  		   \\
			ResNet34 \cite{ResNet} & \textbf{84.15} & 65.37  		& \textbf{74.76} & 37  		   \\
			HRNet-W18 \cite{HRNet} & 83.23 		   & 64.08  		& 73.62    		 & 41  		   \\
			HRNet-W32 \cite{HRNet} & 83.87 		   & \textbf{65.57} & 74.72    		 & 29  		   \\\bottomrule
		\end{tabular}
	}
\end{table*}

The results of the Table \ref{table:Abla:backbone} show that the ResNet34 achieves the best performance in terms of AP and mean KPI. The HRNet-W32 backbone achieves the best performance in terms of mean $\mathcal{J}$, and the VGG16 outperforms in the terms of FPS due to its light-weight architecture but not in terms of accuracy. In summary, the ResNet34 backbone achieves the best trade-off between performance and speed, with a frame rate of 37 FPS. Therefore, we choose ResNet34 as the backbone for the proposed method.

\begin{table*}[!htbp]
	\centering
	\captionsetup{font=small}
	\caption{Single-task and state-off-the-art multi-task versus our proposed multi-task method.}
	\label{table:Abla:obj}
	\resizebox{0.8\linewidth}{!}{
		\tablestyle{12pt}{1.02}
		\begin{tabular}{@{\hskip 2pt}cr|ccc@{\hskip 2pt}}\toprule
			\multicolumn{2}{c|}{Task} & AP $\uparrow$    & Mean $\mathcal{J}\uparrow$ & Runtime (s/frames) $\downarrow$\\\midrule
			\multirow{2}[0]{*}{Single Task Learning} & Detection &   \textbf{84.96}    & - & 0.021 \\
			& Segmentation &   -    & \textbf{66.13} & 0.024 \\\midrule
			\multirow{3}[0]{*}{Multi-Task Learning} & Self-Paced \cite{Kumar2010NIPS} &   83.24 \textcolor{tablered}{\tiny (\textbf{-1.72})}    &  63.94 \textcolor{tablered}{\tiny (\textbf{-2.19})} & 0.031 \\
			& DTP \cite{Guo2018ECCV}   &    83.88 \textcolor{tablered}{\tiny (\textbf{-1.08})}   &  64.28 \textcolor{tablered}{\tiny (\textbf{-1.85})} & 0.027 \\\cmidrule{2-5}
			& \textbf{Ours}  &   \textbf{84.15} \textcolor{tablered}{\tiny (\textbf{-0.81})}   &  \textbf{65.37} \textcolor{tablered}{\tiny (\textbf{-0.76})} & 0.027 \\\bottomrule
		\end{tabular}}
\end{table*}%

\noindent\textbf{Effectiveness of learning strategies.} We evaluate the effectiveness of our overall learning strategy as defined in Eq. \ref{eq:6}. Specifically, we compare the performance of our model, trained simultaneously on both tasks using Eq. \ref{eq:6}, against single-task specific models. These single-task models employ the same architecture outlined in Section \ref{subsec:met:network} but are trained exclusively on a single task. Notably, the single-task models are optimized for detection and segmentation tasks separately, representing their respective optimal training targets. The results of this comparison are presented in Table \ref{table:Abla:obj}.

As shown in Table \ref{table:Abla:obj}, our proposed multi-level dynamic resource prioritization achieves an AP score of 84.15\% and a mean $\mathcal{J}$ score of 65.37\% in the multi-task learning setting. Compared to other multi-task learning strategies, such as Self-Paced \cite{Kumar2010NIPS} and DTP \cite{Guo2018ECCV}, the proposed method demonstrates precision scores for detection and segmentation that are closest to the optimal results achieved by the single-task models. Furthermore, our method achieves both detection and segmentation tasks with a runtime of 0.027 seconds per frame, significantly outperforming multiple single-task models, which require a combined runtime of 0.045 seconds (0.021 for detection and 0.024 for segmentation).

In addition, the proposed sample-level prioritization uses performance metrics to weight samples, and a threshold-based parameter $\delta_{t, i}$ is applied to filter out easy samples, thereby accelerating the training process. We have compared the speed of convergence with DTP and other examples of weighting/filtering methods used in our proposed method, as shown in Table \ref{table:Abla:convergence}.

\begin{table*}[!htbp]
	\centering
	\captionsetup{font=small}
	\caption{The speed of convergence comparison with DTP on the \texttt{val} set of the catheter detection and segmentation dataset.}
	\label{table:Abla:convergence}
	\resizebox{0.9\linewidth}{!}{
		\begin{tabular}{@{\hskip 2pt}c|ccccc@{\hskip 2pt}}\toprule
			\multirow{2}{*}{Methods}    & \multirow{2}{*}{DTP [26]} & \multicolumn{4}{c}{\textbf{Ours}} \\\cmidrule(l){3-6} 
			&                      & w/. Soft Assignment & w/. Hard Assignment & w/. $\mathtt{Top}^{K}$ Soft Assignment & w/. $\mathtt{Top}^{K}$   Hard Assignment \\\midrule
			Convergence   Speed (hours) & 6.1                  & 6.2                 & 5.3                 & 6.3                      & \textbf{5.5}\\\bottomrule                      
		\end{tabular}
	}
\end{table*}

The main reason for the rapid convergence of our method is that the easy samples are filtered out using a threshold, allowing the model to focus more effectively on the difficult samples.

\noindent\textbf{Task weight.} Table \ref{table:Abla:weights} reports the performance of the multi-task total loss function \ref{eq:6} with respect to the different weights $\kappa_{\texttt{d}}$ and $\kappa_{\texttt{s}}$. We compared the constant, variable weighting strategy with our proposed methods.

\begin{table*}[!htbp]
	\centering
	\captionsetup{font=small}
	\caption{Ablation study of the task weight. The symbol `$\equiv$' indicates always equal to the constant value.}
	\label{table:Abla:weights}
	\resizebox{0.65\linewidth}{!}{
		\tablestyle{12pt}{1.02}
	\begin{tabular}{@{\hskip 2pt}c|ccc@{\hskip 2pt}}\toprule
		Weights & AP $\uparrow$   & Mean $\mathcal{J}\uparrow
		$ & Mean KPI $\uparrow$\\\midrule
		$\kappa_{\texttt{d}}\equiv1$, $\kappa_{\texttt{s}}\equiv1$ &   82.86   &	63.45 &	73.16 \\
		$\kappa_{\texttt{d}}\equiv2$, $\kappa_{\texttt{s}}\equiv1$ &   82.94   &	63.39 &	73.17 \\
		$\kappa_{\texttt{d}}\equiv5$, $\kappa_{\texttt{s}}\equiv1$ &   83.57   &	63.22 &	73.40 \\
		$\kappa_{\texttt{d}}\equiv10$, $\kappa_{\texttt{s}}\equiv1$ &    83.78   &	60.87 &	72.33\\
		$\kappa_{\texttt{d}}\equiv1$, $\kappa_{\texttt{s}}\equiv2$ &   82.77   &	62.06 &	72.42 \\
		$\kappa_{\texttt{d}}\equiv1$, $\kappa_{\texttt{s}}\equiv5$ &   82.74   &	62.11 &	72.43 \\
		$\kappa_{\texttt{d}}\equiv1$, $\kappa_{\texttt{s}}\equiv10$ &    82.44   & 62.39 &	72.42\\
		GradNorm \cite{Chen2018ICML} &   78.46    &	56.22 &	67.34\\\midrule
		\textbf{Ours}  &   \textbf{84.15}    &  \textbf{65.37} & \textbf{74.76} \\\bottomrule
	\end{tabular}%
}
\end{table*}%

For the constant weighting strategy, we directly set $\kappa_{\texttt{d}}=\{$1, 2, 5, 10, 1, 1, 1$\}$ and $\kappa_{\texttt{d}}=\{$1, 1, 1, 1, 2, 5, 10$\}$ respectively. Adopting the constant weight loss function does not outperform our multi-level adaptive weighting strategy in terms of mean KPI. In multi-task learning with variable weighting schemes, GradNorm \cite{Chen2018ICML} which automatically adjusts weights for loss according to gradient achieves a \textbf{7.46}\% lower mean KPI compared to our method.

\noindent\textbf{Resource prioritization metrics.} We evaluate the different metrics for KPIs $\kappa$ to define sample and task difficulty $\mathcal{D}$ on the \texttt{val} set to get a better impression of the performance. Table \ref{table:Abla:metrics} shows the results of the detection and segmentation metrics set by the MAE, RMSE, Focal Loss, and IoU, Dice, CE Loss.

\begin{table*}[!htbp]
	\centering
	\captionsetup{font=small}
	\caption{Ablation study of the resource prioritization metrics.}
	\label{table:Abla:metrics}
	\resizebox{0.85\linewidth}{!}{
		\tablestyle{12pt}{1.02}
		\begin{tabular}{@{\hskip 2pt}cc|ccc@{\hskip 2pt}}\toprule
			\multicolumn{2}{c|}{Metrics} & \multirow{2}[0]{*}{AP $\uparrow$} & \multirow{2}[0]{*}{Mean $\mathcal{J}\uparrow$} & \multirow{2}[0]{*}{Mean KPI $\uparrow$}\\
			Detection & Segmentation &       &  &\\\midrule
			MAE   & IoU   &   \textbf{84.15}    & \textbf{65.37} & \textbf{74.76}\\
			MAE   & Dice  &    84.08   & 65.22 & 74.65\\
			RMSE  & IoU   &    83.92   &  65.26 & 74.59\\
			RMSE  & Dice  &    84.28   &  64.88 & 74.58\\
			Focal Loss & CE    &    77.65   & 59.70 & 68.68 \\\bottomrule
		\end{tabular}%
	}
\end{table*}%

We notice that the detection and segmentation metrics defined by the Focal Loss and the CE Loss yield the worst performance compared to the definition by the precision metric. An improper metric is problematic, and the loss cannot reveal the difficulty of samples and tasks and guarantee that more learning resources are allocated to difficult tasks rather than easier ones. We also compare different precision metrics for the KPIs. We see that the KPIs based on the precision metrics outperform the loss-based KPIs, and the proposed method can achieve promising performance when we define the detection and segmentation KPIs by MAE and IoU respectively.

\noindent\textbf{Sample prioritization strategies.} Table \ref{table:Abla:order} compares different sample prioritization strategies and shows the importance of sample-level resource prioritization for multi-task learning. We first sort all samples by the KPIs in the training procedure and apply 7 kinds of samples selected methods.

\begin{figure*}[!htbp]
	\begin{center}
		\includegraphics[width=0.9\textwidth]{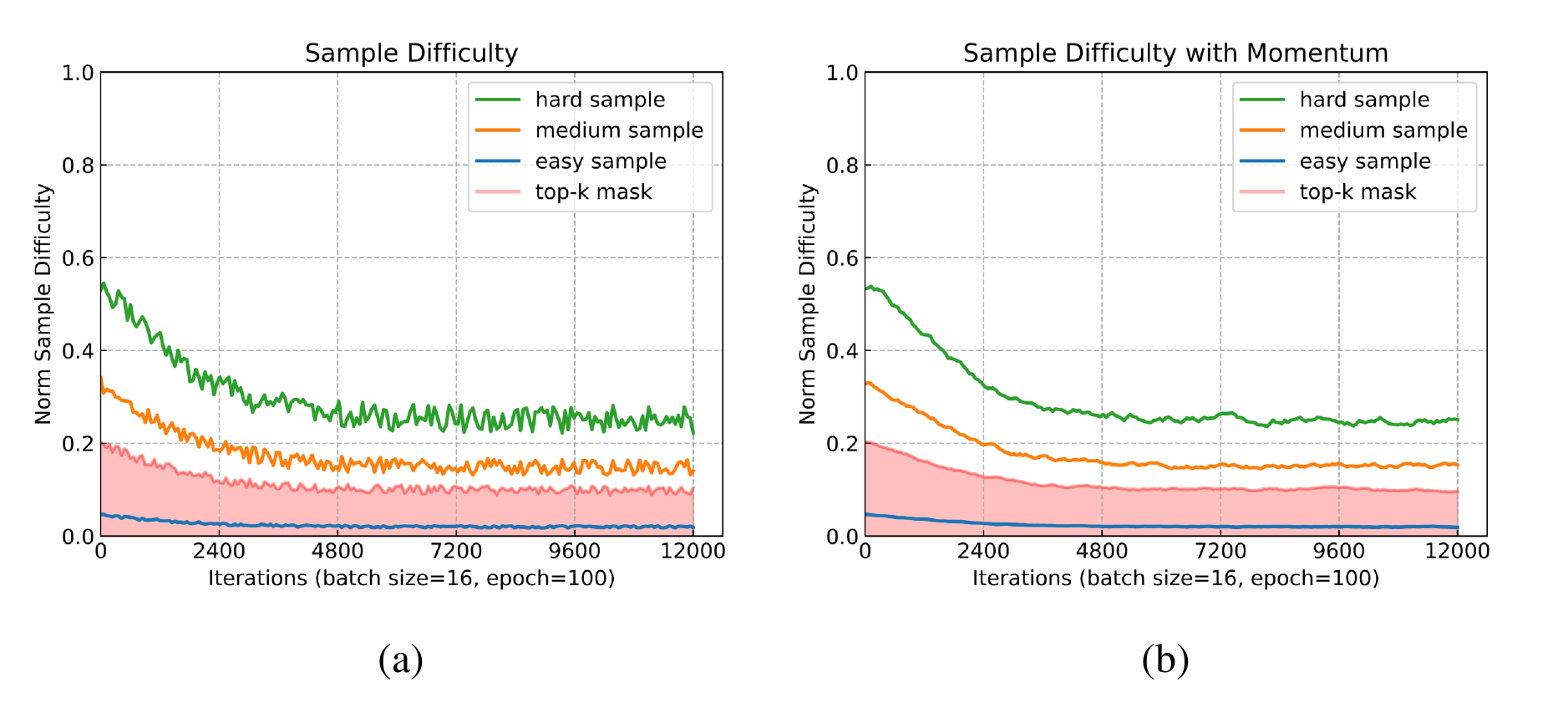}
	\end{center}
	\caption{Difficulty change curve of the selected samples during the training process. The x-axis represents the training iterations, and the y-axis represents the difficulty of the samples. The blue, orange, and green lines represent the easy, medium, and hard samples, respectively. The red mask represents the top-k filtering threshold, which retains 70\% of the samples. (a) The difficulty curve of the samples without momentum update. (b) The difficulty curve of the samples with momentum update.}
	\label{fig:diff_curve}
\end{figure*}

\begin{table*}[!htbp]
	\centering
	\captionsetup{font=small}
	\caption{Ablation study of the sample selection strategy.}
	\label{table:Abla:order}
	\resizebox{0.9\linewidth}{!}{
		\tablestyle{12pt}{1.02}
		\begin{tabular}{@{\hskip 2pt}c|ccc@{\hskip 2pt}} \toprule
		Sample Sort and Selected Strategy & AP $\uparrow$   & Mean $\mathcal{J}\uparrow$ & Mean KPI $\uparrow$ \\\midrule
		Soft Assignment &    81.68   &  61.25 & 71.47 \\
		Hard Assignment ($\delta>$ 0.5) &  83.47     &  63.22 & 73.80 \\
		30\% Hard Samples &     82.92  & 61.08  & 72.00\\
		50\% Hard Samples &    83.11   & 61.46  & 72.29\\
		70\% Hard Samples &   83.19    &  62.85 & 73.02\\
		$\mathtt{Top}^{K}$ Soft Assignment &  84.03     & 64.23  & 74.31\\
		$\mathtt{Top}^{K}$ Hard Assignment &   \textbf{84.15}    &  \textbf{65.37} & \textbf{74.76} \\\bottomrule
	\end{tabular}%
}
\end{table*}%

If we select all of the samples for training and define the weights of each sample by their KPI metrics, such as soft and hard assignment methods, the mean KPI scores reach 71.47\% and 73.80\% respectively. And the training samples are selected with 30\%, 50\%, and 70\% hard samples (defined by KPI) and the final KPI scores are 72.00\%, 72.29\%, and 73.02\%. The optimal sample selection scheme is to dynamically select the number of samples during the training process based on the training iterations using the $\mathtt{Top}^{K}$ algorithm to achieve the optical KPI scores of 74.31\% and 74.76\% respectively.

On the other hand, we randomly selected three samples (\emph{i.e.}, \textbf{easy}, \textbf{medium}, and \textbf{hard} samples) from the training set to visualize their difficulty change curve during training, as shown in the Fig. \ref{fig:diff_curve}. We observe that the difficulty of the samples does not change significantly during the training process, especially when using the momentum update. More fluctuating and unstable samples are filtered out by the top-k algorithm, thus reducing disruptive signals.

\subsection{Runtime Comparison}

To further investigate the computational efficiency of our proposed method, we report the inference time comparisons on the private datasets at 256$\times$256 resolution. We compare our model with the SoTA biomedical image analysis methods that share their codes or include the corresponding experimental results. For the inference time comparison, we run the public code of other methods and our code under the same conditions on the NVIDIA RTX 6000 GPU. The analysis results are summarized in the last column of Table \ref{tab:SoTA}.

As shown in Table \ref{tab:SoTA}, our method achieves 37 FPS, surpassing that of DeepLabV3 \cite{DeepLabV3}, which has the same backbone architecture as our methods. Our model achieves a more favorable accuracy-efficiency trade-off than the existing best method SANet \cite{SANET}, while achieving higher accuracy. In practice, our model yields real-time speed.

\section{Conclusion and Discussions}\label{sec:con}

In this paper, we present a novel convolutional neural network architecture capable of electrode localization using a center heatmap and catheter segmentation. We also propose a novel multi-level dynamic resource prioritization method for multi-task learning, designed to optimize performance across both tasks. Our method dynamically adjusts the weights of both samples and tasks based on their difficulty. Experimental results clearly demonstrates that our method outperforms existing state-of-the-art (SoTA) approaches in both electrode detection and catheter segmentation tasks, using our multi-task learning strategy. Furthermore, our method achieves real-time performance at 37 FPS, simultaneously detecting electrode positions and segmenting the catheter. On the other hand, using separate models for detecting electrode positions and segmenting the catheter may not achieve real-time performance due to the increased computational overhead and doubled inference time. Our framework's output enables the recognition of electrode catheter types \cite{Ma2018MP} based on electrode patterns. The computational cost of post-processing for this recognition is negligible, as it only involves a few dot-product and Euclidean distance calculations \cite{Ma2018MP}. Finally, the computational cost of extracting the centerline from the catheter segmentation results is minimal, as it only involves removing neighbouring pixels and generating a one-pixel-wide skeleton \cite{Ma2018MP}.

By utilizing the outputs from our framework, we can develop numerous clinical applications for enhanced image guidance in cardiac interventional procedures. For example, it can be used for real-time motion compensation to improve the registration accuracy between the 3D roadmap and live X-ray fluoroscopic images. This can be achieved by computing a motion model using a stationary catheter, such as one positioned inside the coronary sinus \cite{Panayiotou2013}. It also can be used for enhancing the visualization of electrode catheters in 3D transesophageal echocardiography \cite{Ma2013UMB, Wu2015} using both 3D echo image volume and live X-ray fluoroscopic images. Furthermore, our framework could pave the way for an advanced computer vision system in future surgical robots by providing real-time positions of each catheter.

The presence of similar objects or complex backgrounds can degrade the model's performance, which explains why our method performs worse when multiple catheter-like object distractors are present. One such low accuracy case is shown on the last row of Fig. \ref{fig:ods_qual}. This issue could be addressed by incorporating spatial constraints for each target object within the video sequence. In the future, we aim to extend our multi-task learning method to tackle more complex tasks and datasets, such as brain tumor classification, detection and segmentation, as well as melanoma segmentation and detection from skin images.`

\section*{Acknowledgements}

This work was supported by EPSRC UK (EP/X023826/1).

\section*{Declarations}

\noindent\textbf{Conflict of interest} The authors declare that they have no conflict of interest.
\vspace{10pt}

\noindent\textbf{Ethical approval} Approval was obtained from the ethics committee of Guy's and St Thomas' NHS Foundation Trust (Ethics approval number: IRAS 150161).
\vspace{10pt}

\noindent\textbf{Informed consent} Informed consent was obtained from all individual patients included in the study.

\bibliography{mybib}

\end{document}